\documentclass{article}

\usepackage{microtype}
\usepackage{graphicx}
\usepackage{booktabs} 

\usepackage{hyperref}

\usepackage[accepted]{icml2025}

\usepackage{amsmath}
\usepackage{amssymb}
\usepackage{mathtools}
\usepackage{amsthm}

\usepackage[capitalize,noabbrev]{cleveref}

\usepackage{multirow}
\usepackage{subcaption}

\usepackage{url}

\theoremstyle{plain}

\theoremstyle{definition}

\theoremstyle{remark}

\usepackage[textsize=tiny]{todonotes}

\icmltitlerunning{Perspective Dial: Measuring Text Perspective and A Prompt-Engineering Method for Controlling Large-Language Model Output Perspective}

\begin{document}

\twocolumn[
\icmltitle{Perspective Dial: Measuring Perspective of Text and Guiding LLM Outputs}



\icmlsetsymbol{equal}{*}

\begin{icmlauthorlist}
\icmlauthor{Taejin Kim}{equal,yyy}
\icmlauthor{Siun-Chuon Mau}{equal,yyy}
\icmlauthor{Konrad Vesey}{yyy}
\end{icmlauthorlist}

\icmlaffiliation{yyy}{CACI International Inc., New Jersey, USA}

\icmlcorrespondingauthor{Taejin Kim}{taejin.kim@caci.com}
\icmlcorrespondingauthor{Siun-Chuon Mau}{siunmau@alumni.princeton.edu}

\icmlkeywords{Large Language Models, Prompt Engineering}

\vskip 0.3in
]



\printAffiliationsAndNotice{\icmlEqualContribution} 


\begin{abstract}
Large language models (LLMs) are used in a variety of mission-critical roles. Due to the rapidly developing nature of LLMs, there is a lack of quantifiable understanding of the bias and perspective associated with LLM output. 
Inspired by this need, this paper considers the broader issue of perspective or viewpoint of general text and perspective control of large-language model (LLM) output.
Perspective-Dial consists of two main components: a (1) metric space, dubbed Perspective Space, that enables quantitative measurements of different perspectives regarding a topic, and the use of (2) Systematic Prompt Engineering that utilizes greedy-coordinate descent to control LLM output perspective based on measurement feedback from the Perspective Space.
The empirical nature of the approach allows progress to side step a principled understanding of perspective or bias -- effectively quantifying and adjusting outputs for a variety of topics.
Potential applications include detection, tracking and mitigation of LLM bias, narrative detection, sense making and tracking in public discource, and debate bot advocating given perspective.
\end{abstract}


\section{Introduction}
\label{sec:intro}

The field of Natural Language Processing (NLP) has undergone a paradigm shift with the advent of Large Language Models (LLMs). Representative LLMs include GPT-4 \cite{openai2024gpt4technicalreport} and Llama \cite{touvron2023llamaopenefficientfoundation} that
are trained to model and generate human-like text via learning from vast corpora of text data. Such models with the capability of performing tasks such as, text summarization \cite{basyal2023textsummarizationusinglarge}, sentiment analysis \cite{zhang2018deeplearningsentimentanalysis}, and automated question-answering have become integral across diverse sectors including education \cite{BERNABEI2023100172}, healthcare \cite{Clusmann2023}, and mission-critical roles \cite{esposito2024largelanguagemodelsmissioncritical}.

However, this proliferation of LLMs in high-stakes environments raises important concerns about the potential presence of embedded, yet undetected, biases and perspectives within these models. 
Biases and associated perspectives manifest as gender, racial, cultural, and socio-economic stereotypes \cite{guo2024biaslargelanguagemodels, muslim-bias}. 
Alignment of LLMs has been introduced as a remedy for such biases --- where various training and fine-tuning is used to promote outputs that are considered less biased and more ethical \cite{shen2023largelanguagemodelalignment}.
However, such alignment has been found to reduce the predictive power of LLMs \cite{bekbayev2023poisonalignment}, or even generate output biased in an unintended direction, as with Google Gemini in 2024 \cite{gemini_debacle}.
The existence of such biases can have far-reaching consequences for users who rely on LLM-generated content, potentially perpetuating or amplifying societal prejudices and misconceptions. As LLMs continue to shape public discourse and influence decision-making processes, it becomes crucial to not only identify and mitigate bias but also to understand and quantify the broader concept of text "perspective".
Then apply perspective control at the output of LLM (System Prompt Engineering) to mitigate undesirable biases. As explained later, perspective is define empirically and implicitly by training data and is semantically similar to viewpoint. In our view, perspective is related to bias as follows: a perspective is biased if it is undesirable.

\begin{figure}[t!]
\begin{center}
\centerline{\includegraphics[width=0.95\columnwidth]{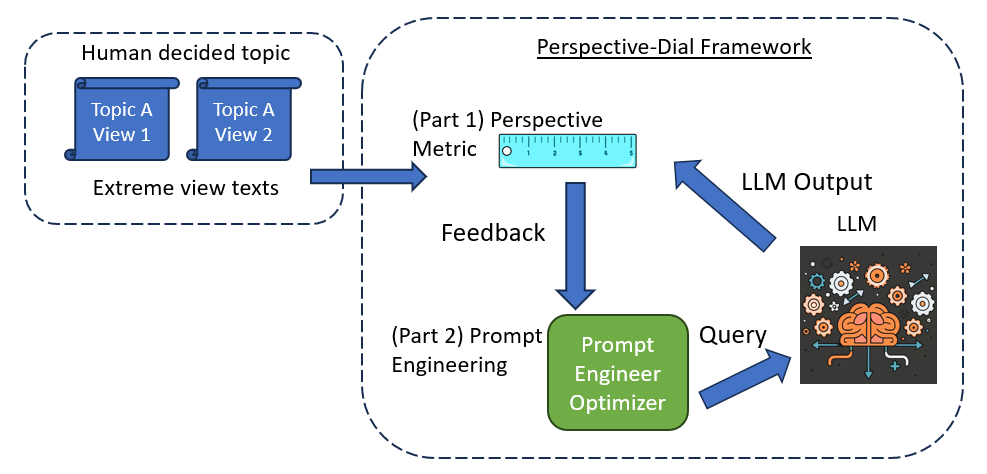}}
    \vskip -0.1in
\caption{Overview of Perspective-Dial pipeline. Given example texts of contrasting perspectives on the same topic, (i) a perspective metric is developed. Based on this perspective metric, (ii) prompt-engineering is performed to tune the perspective of an output of an LLM regarding the relevant topic.}
\label{fig:pdial_overview}
\end{center}
\vskip -0.3in
\end{figure} 

\textbf{Perspective Dial.} This paper proposes Perspective-Dial, a method to quantity, measure and control perspective, with applications to detect, measure and mitigate biases. Other applications include narrative detection, sense making and tracking in public discourse, and debate bot advocating given perspective.
The components are outlined in Figure \ref{fig:pdial_overview}. Perspective-Dial consists of two main components: a (1) Perspective Space, a metric space, inherited from the embedding space of a language model, that allows the quantitative measurements of different perspectives regarding a topic trained via contrastive learning, and (2) Systematic Prompt Engineering that optimizes user prompts to direct the LLM output perspective towards a user prescribed perspective based on output-perspective measurement feedback from the Perspective Space.

\textbf{Challenges and Contributions.} The primary challenge in implementing the proposed Perspective-Dial pipeline lies in the fact that the notion of perspective with relation to LLMs has not been thourougly examined previously. This challenge is compounded by the fact that there is no consesus on the notion of perspective, as well as data sets that can be used to evaluate the performance of Perspective-Dial.

In this paper, our main contributions are the following:
\begin{itemize}
    
	\item We are the first, to the best of our knowledge, to empirically quantify the notion of perspective and apply it to control LLM-output perspective. Using an empirical approach allows us to side step the difficult task of defining perspective from first principle. A contrastive learning technique that utilizes a BERT-based siamese architecture is used to develop the perspective metric. Although bias and sentiment are related to perspective, we view perspective as more general and complex. While we do not plan to delineate exactly how the three concepts are related due to our empirical approach, we believe that our approach is applicable to bias detection and sentiment analysis (future work).

	\item Based on the developed perspective metric, we propose a realistically deployable prompt-engineering scheme to iteratively adjust the perspective of LLM output, toward a user prescribed perspective in the Perspective Space, given a general query. 
    The control of LLM output is agnostic to the Perspective Space, since the LLM backing the Perspective Space is unrelated to the LLM generating text output.
	
	\item To assess the performance of Perspective-Dial with regard to measuring perspective and performing LLM output perspective control, we conduct initial evaluations of both components. We discuss the impact of data size required to train the perspective space, and show that Perspective-Dial can be effectively implemented with a limited labeled data set.
\end{itemize}

The remainder of this paper is organized as follows. 
Section \ref{sec:related} contrasts Perspective-Dial with related works. 
Section \ref{sec:metric} discusses how to define text perspective empirically, and how a perspective metric is developed to measure it. 
Section \ref{sec:prompt_eng} introduces a prompt-engineering scheme that utilizes the trained perspective metric to steer LLM output towards a desired perspective.
We conclude in Section \ref{sec:conc}.


\section{Related Works} \label{sec:related}


\textbf{Bias Detection.}
Bias detection for LLMs is performed by feeding an LLM template-based queries and recording the changes in output \cite{Salinas_2023, stereoset, bold-dataset}. For example, a template is used to observe how an LLM changes its response when asked to recommend a job for a friend of a certain gender and nationality. An unbiased LLM does not change its response given changes in attributes in the query; in practice LLMs often produce different responses based on stereotypes associated with the attributes, indicating a presence of bias in LLMs in general. It is important to note that the detection of biased performed with such tools is different from quantifying perspective Furthermore, bias-measurement tools often are sensitive to additional details of template deployment and unreliable for longer texts \cite{seshadri2022quantifyingsocialbiasesusing}.

\textbf{Sentiment Analysis.}
Sentiment analysis is the process of analyzing text to determine if the emotional tone of the message is positive, negative, or neutral. Sentiment analysis \cite{AWSsentiment} is most often performed by a rule-based method that looks for key words indicating an emotion, or machine learning techniques such as the utilization of language models \cite{zhang-etal-2024-sentiment}. Although text perspective depends on its sentiment, sentiment alone is insufficient to determine perspective. For example, in the context of two rival sports team, a sports fan having positive sentiment towards their own team and negative sentiment towards their rival team represents the same perspective and sentiment analysis can only  report opposite sentiments. Sentiment analysis is similar insufficient for detecting the perspective of a longer piece of text containing multiple intertwined sentiments while nonetheless having a consistent perspective (like having both positive sentiment towards own team and negative sentiment towards rival team).

\textbf{LLM Alignment and Prompt-Engineering.}
Alignment of large language models (LLMs) has emerged as a critical research area, focusing on ensuring that model outputs align with human intent and values. Alignment strategies generally fall into two categories: training-time alignment and post-training methods such as prompt engineering. Training-based approaches often involve supervised fine-tuning on curated datasets that reflect human preferences, followed by reinforcement learning from human feedback (RLHF) to further optimize behavior (Ouyang et al., 2022). RLHF typically uses a reward model trained on comparisons of model responses, which is then used to fine-tune the LLM using reinforcement learning algorithms like Proximal Policy Optimization (PPO) (Christiano et al., 2017). Prompt engineering, on the other hand, leverages careful design of input prompts to elicit desirable outputs from models without altering the model weights (Reynolds \& McDonell, 2021). Recent methods such as chain-of-thought prompting (Wei et al., 2022) and instruction tuning (Sanh et al., 2022) have shown that prompt-based strategies can significantly improve model alignment on complex tasks. These approaches often complement training-based alignment methods, and hybrid pipelines combining both are increasingly common in practice. 

Prompt engineering is the practice of designing inputs for LLMs as to produce optimal or desired outputs. As seen in \cite{Liu2023JailbreakingCV}, prompt engineering can be used to bypass the safeguards installed into services such as ChatGPT to produce radically different output from expected behavior. As discussed in a later section, the added prompt to the input may be optimized in an iterative manner based on a desired loss function to achieve desired output from an LLM \cite{Zou2023UniversalAT}. Perspective-Dial aims to use such an iterative optimization procedure with a loss function based on measurement feedbacks from the Perspective Space.


\section{Measuring Perspective of Text} \label{sec:metric}

The goal of the first innovation – the quantifiable metric space for perspectives of texts – is to move beyond the existing metrics of bias detection within LLMs and sentiment analysis of shorter texts. The perspective metric space is derived from the embedding of texts generated by existing LLMs. 
Utilizing the embeddings generated by LLMs offers the following advantages:

\begin{itemize}
\item The text to which perspective is measured does not have to be in a specified format or length, as LLM embeddings can project texts of varying length into the embedding space with the same format.
\item It is possible to leverage the knowledge the LLM already possesses regarding sentiment of specific words before fine-tuning the embedding space to analyze the perspective of a specific topic.
\end{itemize}

The embedding space of LLMs, explained in \cite{devlin-etal-2019-bert} regarding the BERT encodings, is already used to compare pieces of text with the purpose of language translation, sentiment analysis and image captioning \cite{zhang2020bertscore}. We propose that similar embeddings can be fine-tuned to capture difference in perspective for different pieces of text.

\textbf{Contrastive Learning with Siamese Networks.} A Siamese network is a type of neural network architecture that consists of two or more identical subnetworks working in parallel. These subnetworks share the same parameters and weights, which are updated simultaneously during training. The key feature of Siamese networks is their ability to learn similarities or differences between pairs of inputs.
In contrastive learning, Siamese networks are used to learn representations of data in a way that similar samples are mapped close together in the embedding space, while dissimilar samples are pushed apart. This is achieved by training the network on pairs of samples, where each pair is labeled as either similar or dissimilar \cite{invariant_mapping}.

\begin{figure}[t!]
\begin{center}
\centerline{\includegraphics[width=0.55\columnwidth]{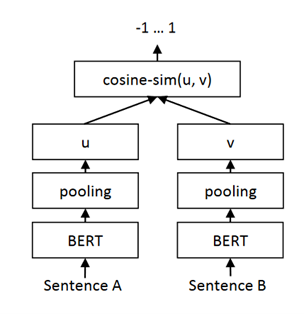}}
    \vskip -0.1in
\caption{Contrastive learning via siamese network used via BertScore in \cite{zhang2020bertscore}.}
\label{fig:siamese}
\end{center}
\vskip -0.3in
\end{figure} 

\textbf{Preparation of Data}
Texts, each holding a specific perspective, are labeled in pairs with other texts in the data corpus. Texts of similar perspectives will share a positive label (1), while texts of dissimilar perspectives will have a negative label (0).
The loss function that is used for the following experiments is:
\begin{itemize}
\item Cosine Similarity Loss:  here the label is continuous in [0,1]
\item Contrastive Loss: The loss function is the Euclidean distance between two embeddings, while the label is binary in [0,1] (1 if 2 texts are similar perspectives, 0 if 2 texts are dissimilar perspectives). A minimum margin value is set between texts that are dissimilar to push the embedding of two texts’ Euclidean distance by at least that margin during training. 
\end{itemize}

Two data sets are manually collected and labeled for training the perspective space:
\begin{itemize}
	\item \textbf{Soccer Example.} Data points manually gathered from Quora. Perspective clusters include Pro-Real Madrid, Pro-Barcelona, and neutrals who are fans of neither team.
	\item \textbf{US Election 2024 Example.} Opinion pieces gathered around the American presidential election of 2024. Perspective clusters include right-leaning, left-leaning, and center.
\end{itemize}

For each of the data sets, opposite clusters were labeled to share a similarity value of 0, while the center/neutral cluster shared a label of 0.35 with each of the poles. Texts within the same cluster share a label of 1. It is important to note that the labels assigned to each text was based of the authors' interpretation of the text.

\begin{table*}[t]
    \centering
    \begin{minipage}[t]{0.48\textwidth}
        \centering
        \caption{Performance of perspective space measurements on the soccer example data set.}
        \label{tbl:soccer_perspective_space}
        \begin{tabular}{lll}
        \hline
        \multicolumn{1}{l|}{Train Set}  & \multicolumn{1}{l|}{Pre-Train}   & Post-Train  \\ \hline
        Pro-Madrid                      & (Test)                           &             \\ \hline
        \multicolumn{1}{l|}{Pro-Madrid} & \multicolumn{1}{l|}{\textbf{0.61 (0.14)}} & \textbf{0.79 (0.25)} \\
        \multicolumn{1}{l|}{Neutral}    & \multicolumn{1}{l|}{0.57 (0.11)} & 0.45 (0.06) \\
        \multicolumn{1}{l|}{Pro-Barca}  & \multicolumn{1}{l|}{0.61 (0.10)} & 0.32 (0.31) \\ \hline
        Neutral                         & (Test)                           &             \\ \hline
        \multicolumn{1}{l|}{Pro-Madrid} & \multicolumn{1}{l|}{0.61 (0.12)} & 0.47 (0.21) \\
        \multicolumn{1}{l|}{Neutral}    & \multicolumn{1}{l|}{\textbf{0.66 (0.12)}} & \textbf{0.91 (0.17)} \\
        \multicolumn{1}{l|}{Pro-Barca}  & \multicolumn{1}{l|}{0.60 (0.10)} & 0.30 (0.06) \\ \hline
        Pro-Barca                       & (Test)                           &             \\ \hline
        \multicolumn{1}{l|}{Pro-Madrid} & \multicolumn{1}{l|}{0.63 (0.09)} & 0.01 (0.06) \\
        \multicolumn{1}{l|}{Neutral}    & \multicolumn{1}{l|}{0.58 (0.09)} & 0.34 (0.02) \\
        \multicolumn{1}{l|}{Pro-Barca}  & \multicolumn{1}{l|}{\textbf{0.71 (0.11)}} & \textbf{0.98 (0.02)} \\ \hline
        \end{tabular}
    \end{minipage}
    \hfill
    \begin{minipage}[t]{0.48\textwidth}
        \centering
        \caption{Performance of perspective space measurements on the election example data set.}
		\label{tbl:election_perspective_space}
        \begin{tabular}{lll}
		\hline
		\multicolumn{1}{l|}{Train Set}  & \multicolumn{1}{l|}{Pre-Train}   & Post-Train  \\ \hline
		Left Lean                       & (Test)                           &             \\ \hline
		\multicolumn{1}{l|}{Left Lean}  & \multicolumn{1}{l|}{\textbf{0.45 (0.18)}} & \textbf{0.53 (0.33)} \\
		\multicolumn{1}{l|}{Center}     & \multicolumn{1}{l|}{0.37 (0.10)} & 0.41 (0.06) \\
		\multicolumn{1}{l|}{Right Lean} & \multicolumn{1}{l|}{0.43 (0.16)} & 0.62 (0.34) \\ \hline
		Center                          & (Test)                           &             \\ \hline
		\multicolumn{1}{l|}{Left Lean}  & \multicolumn{1}{l|}{0.50 (0.14)} & 0.44 (0.01) \\
		\multicolumn{1}{l|}{Center}     & \multicolumn{1}{l|}{\textbf{0.59 (0.17)}} & \textbf{0.95 (0.00)} \\
		\multicolumn{1}{l|}{Right Lean} & \multicolumn{1}{l|}{0.48 (0.14)} & 0.37 (0.01) \\ \hline
		Right Lean                      & (Test)                           &             \\ \hline
		\multicolumn{1}{l|}{Left Lean}  & \multicolumn{1}{l|}{0.38 (0.18)} & 0.44 (0.28) \\
		\multicolumn{1}{l|}{Center}     & \multicolumn{1}{l|}{0.30 (0.11)} & 0.47 (0.05) \\
		\multicolumn{1}{l|}{Right Lean} & \multicolumn{1}{l|}{\textbf{0.37 (0.18)}} & \textbf{0.67 (0.34)} \\ \hline
		\end{tabular}

    \end{minipage}
\end{table*}

For the soccer example as seen in Table \ref{tbl:soccer_perspective_space}, it is observed before any training occurs, that there is only mild differentiation between clusters. For example, taking a test text from a pro-Madrid perspective and comparing it to the training text of all clusters, it is seen that the text has equal cosine similarity values with the pro-Madrid cluster and the pro-Barcelona cluster. Once contrastive learning is performed, the measurements of perspective for train-set texts and test-set texts become much more distinguished. After training, a pro-Madrid text most heavily aligns with the pro-Madrid train set, while less so with the neutral cluster and pro-Barcelona cluster respectively. The converse is true for the pro-Barcelona texts, which align with the pro-Barcelona train set most, and gradually decreases against the neutral train set and pro-Madrid train set respectively. Thus, it is seen for the soccer example that the perspective space successfully measures perspective between supports of the two different teams.

For the election example as seen in Table \ref{tbl:election_perspective_space}, similar to the soccer example, it is observed before any training occurs, that there is only mild differentiation between clusters. For example, taking a test text from a left learning perspective and comparing it to the training text of all clusters, it is seen that the text has very close cosine similarity values with the left leaning train set and the right leaning train set. Once contrastive learning is performed, the measurements of perspective for train-set texts become much more distinguished. When comparing texts from the train set of each cluster to each of the train set, the capability of recognizing perspective is clear, as left leaning texts correlate highly with the left leaning train set and vice versa. However, when analyzing the test set, it is seen that perspective is only correctly identified for right leaning and neutral texts. The left leaning test set is more closely aligned with the right leaning train set. Potential improvements to these results include the introduction of more training texts that cover the full scope of each perspective.


\begin{figure}[t!]
\begin{center}
\centerline{\includegraphics[width=0.95\columnwidth]{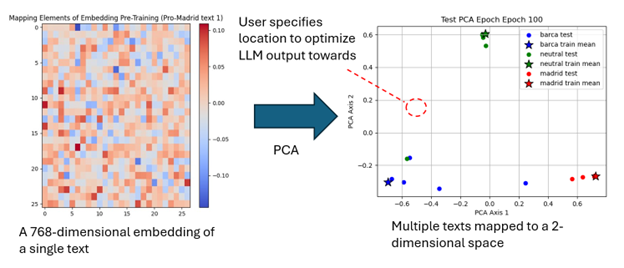}}
    \vskip -0.1in
\caption{Utilization of PCA to compress embedding space to lower dimensions. A user of Perspective-Dial can target for LLM perspective control.}
\label{fig:pca_overview}
\end{center}
\vskip -0.3in
\end{figure} 

\begin{figure}[t!]
\begin{center}
\centerline{\includegraphics[width=0.95\columnwidth]{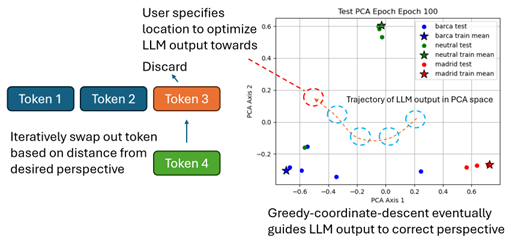}}
    \vskip -0.1in
\caption{Greedy-coordinate-decent used to perform prompt-engineering on LLM output. LLM output gradually shifts towards desired pre-set perspective, as denoted by perspective space.}
\label{fig:gcd_overview}
\end{center}
\vskip -0.3in
\end{figure} 

\begin{figure*}[t]
    \centering
    \begin{subfigure}[t]{0.49\textwidth}
    \includegraphics[width=0.8\textwidth]{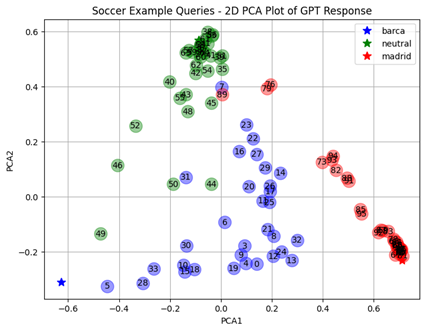}
    \caption{Spread of perspectives for a brute-force search across queries fed into GPT-4 for the soccer example. Each color represents the base-query, asking the LLM to generate text as a fan of Real Madrid, FC Barcelona, or neutral.}
    \label{fig:opt_soccer}
    \end{subfigure}
    \hspace{0.03cm}
    \begin{subfigure}[t]{0.49\textwidth}
    \includegraphics[width=0.8\textwidth]{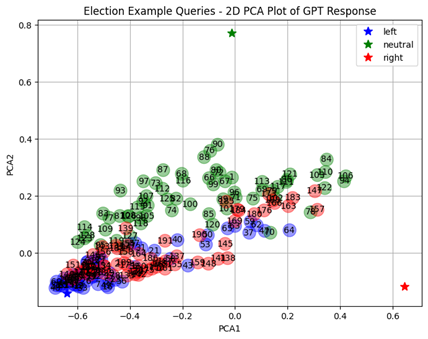}
    \caption{Spread of perspectives for a brute-force search across queries fed into GPT-4 for the US election example. Each color represents the base-query, asking the LLM to generate text as leaning right, left, or neutral.}
    \label{fig:opt_politics}
    \end{subfigure}
  \caption{Greedy-coordinate-descent approach using the perspective metric.}
  \label{fig:group_plot_1}
\end{figure*}

\section{Optimization-Based Prompt Engineering} \label{sec:prompt_eng}

After the perspective space is developed, a user can specify a specific point in the perspective space to optimize LLM output towards. For the sake of simplicity, the embedding space of dimensionality of 768 is reduced to a dimensionality of 2 for user specification via principal component analysis (PCA).
Within the PCA space, a user can now specify a point in 2 dimensions within the perspective space. This procedure is shown in Figure \ref{fig:pca_overview}. Once the desired perspective is specified, the user can optimize via greedy-coordinate-descent. Here, specific tokens are iteratively swapped out and queried to the LLM. Specific tokens that push the LLM output towards the desired perspective are kept in the query, whilst low performing tokens are swapped out. The loss function between the LLM output and the desired perspective in the perspective space is depicted as the L2 norm between the two points in the 2-dimensional PCA space.

Within the PCA space, a user can now specify a point in 2 dimensions within the perspective space. Once the desired perspective is specified, the user can optimize via greedy-coordinate-descent. Here, specific tokens are iteratively swapped out and queried to the LLM. Specific tokens that push the LLM output towards the desired perspective are kept in the query, whilst low performing tokens are swapped out. The loss function between the LLM output and the desired perspective in the perspective space is depicted as the L2 norm between the two points in the 2-dimensional PCA space.

Initial experiments are performed on the aforementioned soccer and election examples data sets. Instead of swapping out individual tokens, each query consists of a [base phrase] + [combination of additional phrases], where the additional phrases are swapped out to perform a simulation of greedy-coordinate-descent (GCD). A visual representation of GCD is shown in Figure \ref{fig:gcd_overview}.

It is shown for the soccer example that when various inputs are passed to the LLM to induce a specific perspective, different combinations of phrases shift the perspective in the trained embedding space. Below are specific outputs of the LLM that are returned by brute-force optimization for a desired perspective.

It is shown for the election example that when various inputs are passed to the LLM to induce a specific perspective, different combinations of phrases shift the perspective in the trained embedding space. Below are specific outputs of the LLM that are returned by brute-force optimization for a desired perspective.

For each data set, three optimization experiments occur – each optimizing for one of the three perspectives represented by the cluster labels (e.g., Pro-Madrid, Pro-Barcelona, and Neutral). Through a brute-force search of all combinations of base phrases and additional phrases, it is shown for both data sets, the correct phrase is returned.


\section{Conclusion} \label{sec:conc}
We introduce Perspective-Dial, a novel approach for quantifying and controlling the perspective of text generated by large language models. By developing a Perspective Space metric and employing systematic prompt engineering, we have demonstrated a method to measure and adjust the viewpoint expressed in LLM outputs. Our empirical approach sidesteps the need for a formal definition of perspective, instead relying on contrastive learning to capture nuanced differences in textual perspectives. The potential applications of Perspective-Dial are far-reaching, from mitigating unintended biases in AI-generated content to enabling more nuanced analysis of public discourse and facilitating perspective-specific debate bots. While our initial evaluations show promise, further research is needed to fully explore the capabilities and limitations of this approach across diverse topics and use cases. As LLMs continue to play an increasingly central role in shaping information and decision-making processes, tools like Perspective-Dial will be crucial for ensuring these powerful systems can be deployed responsibly and effectively.

\bibliography{references}
\bibliographystyle{icml2025}



\end{document}